\documentclass[11pt,a4paper]{article}
\usepackage{times,latexsym}
\usepackage{url}
\usepackage[T1]{fontenc}

\usepackage{graphicx}
\usepackage{color}
\usepackage{xcolor}
\usepackage{booktabs}
\usepackage{multirow}
\usepackage{hyperref}
\usepackage{array}
\usepackage{amsmath}
\usepackage{amsfonts, amssymb}
\usepackage{colortbl}
\usepackage{subfigure}
\usepackage{enumitem}

\usepackage{amsmath}        
\usepackage{algorithm}      
\usepackage{algorithmic}    

\definecolor{purple}{RGB}{128, 0, 128}
\definecolor{LightRed}{rgb}{1,0.92,0.92}
\definecolor{LightOrange}{rgb}{1,0.95,0.88}
\definecolor{LightYellow}{rgb}{1.0,1.0,0.84}
\definecolor{LightGreen}{rgb}{0.9,1.0,0.88}
\definecolor{LightCyan}{rgb}{0.9,1,1}
\definecolor{LightBlue}{rgb}{0.9,0.94,1}
\definecolor{LightIndigo}{rgb}{0.92,0.9,1}
\definecolor{LightMagenta}{rgb}{0.96,0.86,1}
\definecolor{DirtyWhite}{rgb}{0.96,0.96,0.96}

\DeclareSymbolFont{extraup}{U}{zavm}{m}{n}
\DeclareMathSymbol{\varheart}{\mathalpha}{extraup}{86}
\DeclareMathSymbol{\vardiamond}{\mathalpha}{extraup}{87}
\DeclareMathSymbol{\varclubsuit}{\mathalpha}{extraup}{88}

\newcolumntype{L}[1]{>{\raggedright\arraybackslash}p{#1}}
\newcolumntype{C}[1]{>{\centering\arraybackslash}p{#1}}
\newcolumntype{R}[1]{>{\raggedleft\arraybackslash}p{#1}}

\DeclareSymbolFont{extraup}{U}{zavm}{m}{n}
\DeclareMathSymbol{\varheart}{\mathalpha}{extraup}{86}
\DeclareMathSymbol{\vardiamond}{\mathalpha}{extraup}{87}

\usepackage{amsmath}
\usepackage{amssymb}
\usepackage{wasysym}

\DeclareSymbolFont{extraup}{U}{zavm}{m}{n}
\DeclareMathSymbol{\varspadesuit}{\mathalpha}{extraup}{83}
\DeclareMathSymbol{\varheartsuit}{\mathalpha}{extraup}{86}
\DeclareMathSymbol{\vardiamond}{\mathalpha}{extraup}{87}
\DeclareMathSymbol{\varclubsuit}{\mathalpha}{extraup}{88}

\usepackage[acceptedWithA]{tacl2021v1}

\usepackage{xspace,mfirstuc,tabulary}

\newif\iftaclinstructions
\taclinstructionsfalse 
\iftaclinstructions
\renewcommand{\confidential}{}
\renewcommand{\anonsubtext}{(No author info supplied here, for consistency with
TACL-submission anonymization requirements)}
\newcommand{\instr}
\fi

\iftaclpubformat 

\else

\fi

\title{Balanced Training Data Augmentation for\\ Aspect-Based Sentiment Analysis}

\author{
    Junjie Liu$^{{\spadesuit}}$, \hspace{0.1cm}
    Yuanhe Tian$^{\varheart}$, \hspace{0.1cm}
    Yan Song$^{{\spadesuit}}$
    \\
    $^{\spadesuit}$University of Science and Technology of China \hspace{0.1cm}
    $^{\varheart}$University of Washington \\
    $^{\spadesuit}$\texttt{ljj19937347730@mail.ustc.edu.cn} \hspace{0.1cm}
    $^{\varheart}$\texttt{yhtian@uw.edu} \\
    $^{\spadesuit}$\texttt{clksong@gmail.com}
}

\date{}

\begin{document}
\maketitle

\begin{abstract}
Aspect-based sentiment analysis (ABSA) is a crucial fine-grained task in social media scenarios to identify the sentiment polarity of specific aspect terms in a sentence. 
Although many existing studies leverage large language models (LLMs) to perform ABSA due to their strong context understanding capabilities, they still face challenges to learn the context information in the running text because of the short text, as well as the small and unbalanced labeled training data, where most data are labeled with positive sentiment.
Data augmentation (DA) is a feasible strategy for providing richer contextual information, especially when using LLMs to create synthetic training data, but faces challenges in ensuring a high quality of the augmented data.
In this paper, we propose an LLM-based ABSA approach with training data augmentation.
Specifically, an LLM is prompted to generate augmented training data based on the original training data, so as to construct a new training data with larger size and balanced label distributions to better train an ABSA model.
Meanwhile, in order to improve the quality of the augmented data, we propose a reinforcement learning approach to optimize the data augmentation LLM.
Experiment results and further analyses on English benchmark datasets for ABSA demonstrate the effectiveness of our approach, where superior performance is observed over strong baselines and most existing studies.\footnote{The code is available at \url{https://github.com/synlp/RDA-ABSA}.
}
\end{abstract}

\section{Introduction}
Aspect-based sentiment analysis (ABSA) is an important fine-grained sentiment analysis task that focuses on identifying and analyzing specific aspects of the expressed emotions in a sentence, aiming to predict the sentiment polarity of specific aspect terms rather than assess the overall sentiment of the entire sentence.
For example, consider this comment, "\textit{I like this computer except that the screen is a little small.}"
Here, the two aspects of "\textit{computer}" and "\textit{screen}" respectively carry \textit{positive} and \textit{negative} emotions, although the entire sentence tends to express \textit{positive} emotions.
The ABSA task is increasingly widely applied in the real world, especially in user comments and social media, and thus has received much research attention \cite{tang2020dependency,tian2021aspect,chen2022discrete,wang2023reducing,luo2024panosent}.

\begin{figure*}[t]
    \centering
    \includegraphics[width=1.0\linewidth, trim=0 40 0 0]{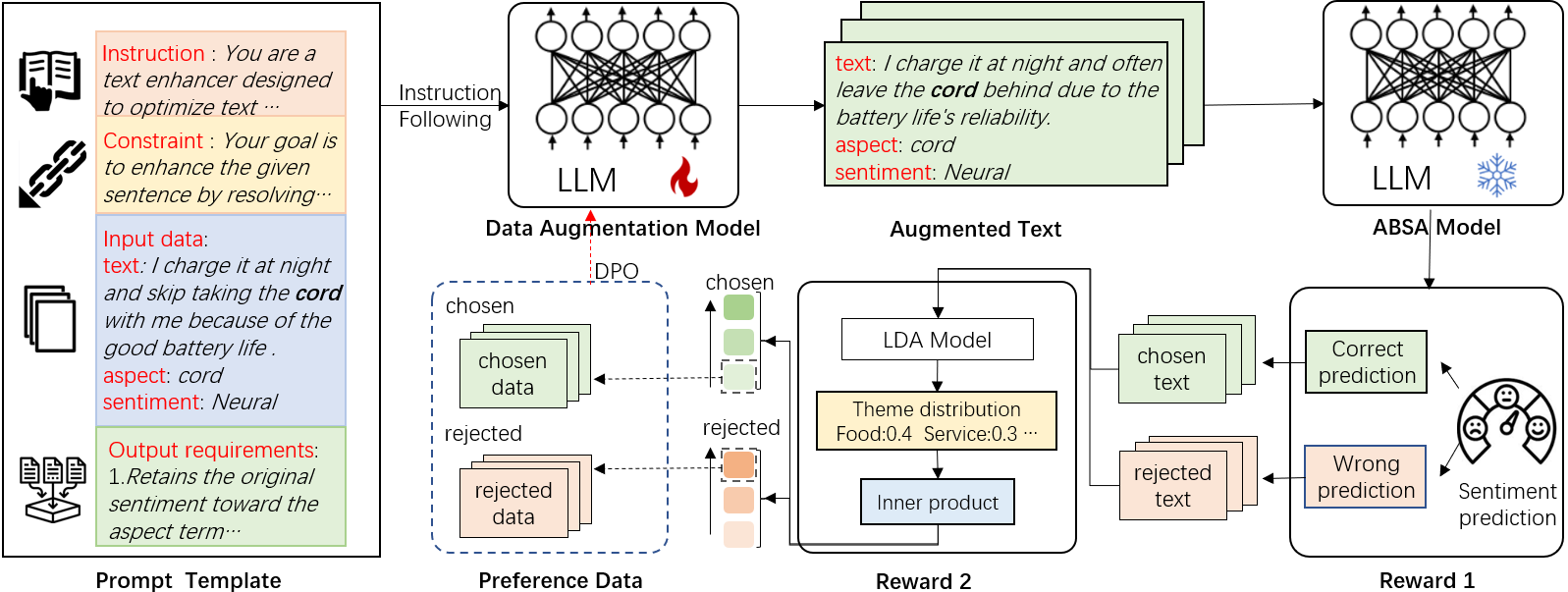}
    \caption{The overall architecture of our approach for ABSA with data augmentation.
    We use a prompt to instruct the data augmentation large language model to generate several augmented instances.
    The augmented instances are then fed into the ABSA model to predict the sentiment, where the output is used to compute the reward to rank the augmented instances.
    A preference dataset is constructed based on reward and is used to perform direct preference optimization (DPO) for optimizing the data augmentation LLM.}
    \label{fig:model}
    \vspace{-0.2cm}
\end{figure*}

Recent studies on ABSA mainly utilize pre-trained language models and large language models (LLMs) as the foundation models \cite{simmering2023large,luo2024panosent,negi2024hybrid,ding2024boosting,zhu2024zzu}.
These language models are trained on a large amount of unlabeled data, thus having a powerful context encoding capability that enables them to analyze the context and accurately predict the sentiment of aspect terms after being trained on labeled data.
Most of the existing data for ABSA comes from social media, and they are generally short and provide limited contextual information.
Meanwhile, the amount of human-annotated data is relatively small, and most data is positive, which results in an unbalanced label distribution that leads an ABSA model to learn the bias in the data and thus is not able to effectively perform the task.
An intuitive idea to address the data issue is to perform data augmentation \cite{zhong2024iterative,lu2025qaie,HELLWIG2025125514} by prompting LLMs to generate augmented training data to help the ABSA model better learn the contextual information in social media text and thus perform the ABSA task better.
However, LLMs face hallucination issues when they are generating texts, making data augmentation challenging for ABSA in terms of the quality of the generated data.
To address this issue, existing studies build scoring mechanisms to assess the quality of the generated texts and filter out the low-quality ones \cite{li2024iterative,Chen2024ReconfidencingLF,AHMED2025128284}.
Most scoring mechanisms focus on the emotional and domain relevance of the texts, or the richness of the vocabulary, where less attention is paid to whether the context information of the enhanced data is rich.

In this paper, we propose a balanced training data augmentation approach for ABSA.
Specifically, we utilize an LLM to generate extra training data based on the original training data, so as to produce a new training data with balanced labeled distribution to train an ABSA model.
To ensure the improvement of the generated text in terms of quality and diversity, we employ reinforcement learning to optimize the data augmentation model, where a reward function is designed to assess the quality of the augmented data from various aspects. 
Experiment results and further analyses on English benchmark datasets for ABSA demonstrate the effectiveness of our approach, where our approach outperforms strong baselines and most existing studies on the datasets.

\section{The Approach}

We propose a data augmentation approach to leverage the augmented training data produced by the LLM to train an ABSA model, so as to improve its performance on the task, where the data augmentation LLM is optimized through reinforcement learning to generate better augmented data.
The overall approach is shown in Figure \ref{fig:model}. 
Specifically, for the sentence $\mathcal{X}$, aspect term $\mathcal{A}$, and sentiment label $\mathcal{Y}^*$ in the original training data, we prompt an LLM to generate augmented text $\mathcal{X}'$ with the same aspect term and the sentiment label.
We combine the original training data and the augmented training data to construct a new training data with balanced sentiment label distribution and train the ABSA model on the data.
To improve the quality of the augmented data, we employ a reinforcement learning process to optimize the LLM, where a reward function is designed to rank the augmented data so as to construct a preference dataset to facilitate direct preference optimization (DPO) \cite{NEURIPS2023_a85b405e}.
In the following text, we firstly describe the ABSA process with our balanced training data augmentation, and then elaborate on the reinforcement learning process to optimize the data augmentation process.

\begin{table*}[t]
    \centering
    \begin{tabular}{p{15cm}}
    \toprule
    \textbf{Prompt:} \\
    \textit{You are a text enhancer designed to optimize text specifically for aspect-based sentiment analysis (ABSA) models by enriching, clarifying, and standardizing content. 
     Your goal is to enhance the given sentence by improving grammar, resolving ambiguities, and inferring missing information, 
     thereby boosting the ABSA model's performance. Given an original sentence, a specific aspect term within that sentence, 
     and the sentiment associated with that aspect term (Positive, Negative, or Neutral), generate a new sentence that:}\\
    \textit{1. Clearly includes the provided aspect term.}\\
    \textit{2. Retains the original sentiment toward the aspect term.}\\
    \textit{3. Is close in length to the original sentence.}\\
    \textit{4. Contains only the enhanced sentence without any additional explanation or irrelevant content.}\\
    \textit{5. Don't annotate (like Here is the enhanced sentence:), do not explain, just output enhanced text.}\\
    \textit{The given sentence, aspect-term, and sentiment are the following:}\\
    \textit{Sentence: ``<sentence>''}\\
    \textit{Aspect: ``<aspect term>''}\\
    \textit{Sentiment: ``<sentiment>''}\\
    \bottomrule
    \end{tabular}
    \vspace{-0.1cm}
    \caption{The prompt used for data augmentation. ``\textit{<sentence>}'', ``\textit{<aspect term>}'', and ``\textit{<sentiment>}'' are placeholders for the sentence, aspect term, and the sentiment polarity label of the training instance.}
    \label{tab:my_label}
    \vspace{-0.2cm}
\end{table*}

\subsection{Balanced Training Data Augmentation}
Our approach firstly utilizes an LLM to perform data augmentation to construct balanced training data and then trains an ABSA model on it.
Specifically, we first design a prompt template to instruct the data augmentation LLM to generate training data. 
The complete prompt template is shown in Table \ref{tab:my_label}. This template constrains and instructs LLM to generate high-quality training data from two aspects, including text normalization and aspect term sentiment consistency.
Among them, text normalization includes instructions to improve the grammar of the input text, eliminating ambiguity and missing inferences; 
aspect term sentiment consistency requires the aspect terms and aspect sentiment of the text generated by LLM to be consistent with the input text and comply with the format requirements of the ABSA training data.
Next, we fill the sentence $\mathcal{X}$, aspect term $\mathcal{A}$, and the gold standard label $\mathcal{Y}^*$ in the original training data into the prompt template.
Afterwards, we feed the prompt into LLM, and LLM generates the enhanced text $\mathcal{X}'$ according to the standard decoding process.
Finally, we combine the enhanced text $\mathcal{X}'$, the aspect term $\mathcal{A}$ of the input text, and the original gold standard label $\mathcal{Y}^*$ into a new training instance $\{\mathcal{X}', \mathcal{A}, \mathcal{Y}^*\}$.

To train an ABSA model with the augmented data, we mix the augmented training data and the original training data.
Considering that the original training data with unbalanced label distribution may introduce label bias to a model trained on it, we perform data augmentation to produce a final training dataset with balanced label distribution and train the ABSA model on it.
Given a sentence and aspect term pair $\{\mathcal{X}, \mathcal{A}\}$ as the model input, the goal of the ABSA model is to produce the sentiment polarity $\widehat{\mathcal{Y}} \in \{\textit{positive},\textit{neutral},\textit{negative}\}$ of the aspect term $\mathcal{A}^*$.
Specifically, we use an LLM as the ABSA model and design a prompt $\mathcal{P}$, i.e., ``\textit{Predict the sentiment of the given aspect in the text.}'', to instruct the LLM to generate the sentiment of the aspect term. 
Then, we feed the prompt $\mathcal{P}$, the sentence, and the aspect-term into the LLM, and the LLM generates the sentiment label $\widehat{\mathcal{Y}}$ following the standard LLM decoding process.
Finally, we compare $\widehat{\mathcal{Y}}$ with the gold standard label to compute the cross-entropy loss $\mathcal{L}$, which is used to update the ABSA model through backpropagation.

\subsection{Reinforced LLM Optimization}
We find that when using prompt templates to instruct LLM in generating augmented data, the diversity of the augmented data and the quality of the text are not stable enough, which obviously has a negative impact on the training of the ABSA models.
Therefore, we propose a reinforcement learning approach to optimize the data augmentation LLM, aiming to enable the model to generate high-quality text.
In doing so, we design a reward function to evaluate the quality of the augmented data and build a preference dataset accordingly for reinforcement learning through DPO.
Our reward function assesses the augmented data from two aspects: whether the aspect term in the augmented data shares the same sentiment with the aspect term in the original data, and whether the topic of the augmented sentence $\mathcal{X}'$ is consistent to the original sentence $\mathcal{X}$.
The details of the two rewards are the following.

For the first reward, we use the sentiment prediction function to evaluate the sentiment consistency between the augmented training data and the original training data.
For each augmented data $\{\mathcal{X}',\mathcal{A},\mathcal{Y}^*\}$ generated by the data augmentation model, we feed the text $\mathcal{X}'$ and the aspect term $\mathcal{A}$ into the untrained ABSA model to predict the sentiment $\widehat{\mathcal{Y}}$.
Then, we compare $\widehat{\mathcal{Y}}$ with the gold standard label $\mathcal{Y}^*$.
If the model prediction matches the gold standard label, it means that the label of the augmented data is more likely to be correct than the case where the prediction $\widehat{\mathcal{Y}}$ does not match $\mathcal{Y}^*$, and thus the augmented data is regarded as high-quality data. 

For the second reward, we utilize the latent Dirichlet allocation (LDA) model to evaluate the topic distribution of the original text and the enhanced text $\{\mathcal{X},\mathcal{X}'\}$ and calculate the relevance.
The differences in the topic representations among different texts reflect their relevance.
The text relevance is computed through the following process.
Given a text input $x$, we firstly utilize the LDA model to generate a topic distribution vector $\mathbf{z}_x$ over $K$ topics, where each element $z_{x,k}$ represents the probability of topic $k$ in document $x$:
\begin{equation}
    \mathbf{z}_x = [z_{x,1}, z_{x,2}, \dots, z_{x,K}]
\end{equation}
where 
\begin{equation}
    z_{x,k} = \frac{\theta_{x,k}}{\sum_{i=1}^{K} \theta_{x,i}}
\end{equation}
with $\theta_{x,k}$ the Dirichlet parameter representing the proportion of topic $k$ in document $D$.
We apply the process to both the original sentence $\mathcal{X}$ and the augmented sentence $\mathcal{X}'$ to obtain their topic representations $\mathbf{z}_{x'}$ and $\mathbf{z}_x$, respectively.
Then, we use the cosine similarity between the two topic representations to measure the relevance between them, which is formulated as
\begin{equation}
    r({\mathcal{X}'}, \mathcal{X}) = \frac{\mathbf{z}_{x'} \cdot \mathbf{z}_x}{\|\mathbf{z}_{x'}\| \cdot \|\mathbf{z}_x\|} 
\end{equation}
where the relevance score $r({\mathcal{X}'}, \mathcal{X})$ ranges from $0$ (which means completely dissimilar) to $1$ (which means identical topic distributions).

\begin{table}[t]
\begin{center}
        \centering
        \begin{tabular}{l l | r | r| r}
        \toprule
        \multicolumn{2}{c|}{\textbf{Dataset}} & \multicolumn{1}{c|}{\textbf{Pos. \#}} & \multicolumn{1}{c|}{\textbf{Neu. \#}} & \multicolumn{1}{c}{\textbf{Neg. \#}} \\
        \midrule
        \multirow{2}{*}{\textbf{LAP14}} & Train & 994 & 464 & 870 \\
        & Test & 341 & 169 & 128 \\
        \midrule
        \multirow{2}{*}{\textbf{REST14}} & Train &2,164 &637 &807 \\
        & Test &728 &196 &182 \\
        \midrule
        \multirow{2}{*}{\textbf{REST15}} & Train &907 &36 &254 \\
        & Test &326 &34 &207 \\
        \midrule
        \multirow{2}{*}{\textbf{REST16}} & Train &1,229 &69 &437 \\
        & Test &469 &30 &114 \\
        \bottomrule
        \end{tabular}
\end{center}
\vskip -0.3cm
\caption{\label{tab: dataset}
        The statistics of the ABSA datasets. We report the number of instances with different sentiment polarities in the training and test sets.
        }
 \vskip -1em    
\end{table}

To build the preference data, for each original training data, we run the LLM for data augmentation multiple times to generate different augmented sentences through sampling.
We use the two rewards to construct the preference data that contains one accepted instance and one rejected instance with the same original training data.
Specifically, we use the first reward to classify the augmented texts into two sets, namely, the chosen set and the rejected set, where the augmented data whose predicted sentiment label matches the gold standard label is categorized into the chosen set and the other data are put into the rejected set.
Then, we select the augmented text with the lowest correlation score as the accepted instance from the chosen set and select the augmented text with the highest correlation score as the rejected instance from the rejected set.
This allows us to select less distinguishing accepted and rejected instances and thus allows the data augmentation LLM to learn from more challenging cases, so as to improve the quality of the generated text.
At the same time, we also take into account the situation where the predicted sentiment of the augmented data is completely consistent with or completely different from the gold standard sentiment label, which leads to an empty chosen set or an empty rejected set.
When the chosen set is empty, we set the enhanced text with the lowest relevance score as the accepted instance, and the enhanced text with the highest relevance score as the rejected instance.
When the rejected set is empty, our settings are exactly the opposite: the enhanced text with the highest relevance score is set as the chosen text $t_c$, while the enhanced text with the lowest relevance score is set as the rejected text $t_r$.
Finally, we use the preference data to optimize the data augmentation LLM through DPO.
Through this process, our data augmentation module learns to generate high quality and diverse training data for improving the performance of the ABSA model.

\section{Experiment Settings}

\subsection{Datasets}
We evaluate different models on four English benchmark datasets for ABSA from different domains. The selected datasets includes LAP14 \cite{pontiki-etal-2014-semeval} that focuses on laptop reviews, and REST14 \cite{pontiki-etal-2014-semeval}, REST15 \cite{pontiki2015semeval}, and REST16 \cite{pontiki2016semeval} that contains restaurant service feedback.
The diversity of these datasets not only reflects the differences in language features and emotional expressions in specific domains but also provides a strict testing environment for the generalization ability of the model. This study strictly adopts the training and test set division officially provided by each dataset\footnote{All datasets do not have an official development set.
}.
Table \ref{tab: dataset} shows the distribution of aspect terms in each dataset based on the sentiment labels of \textit{positive (pos)}, \textit{negative (neg)}, and \textit{neutral (neu)}. 
It is observed from the Table that the data show an unbalanced distribution of sentiment polarity (for example, in REST16, the proportion of positive instances is significantly higher than that of other categories), which generally makes the task more challenging.
This cross-domain differential distribution provides ideal conditions for verifying the robustness of the model.

\subsection{Baseline and Comparing Models}

In the experiments, we compare our approach with two baselines. 
The first is the ``\textbf{base}'' model that only uses the original training data for supervised fine-tuning of the LLM without using any data augmentation. 
The second is the ``\textbf{only DA}'' model that utilizes the augmented data to train the ABSA model, where reinforcement learning is not used to optimize the data augmentation LLM; 
We also compare our approach with the following existing state-of-the-art models for ABSA.

\begin{itemize}[leftmargin=1em, itemsep=0.1ex]
    \item \textbf{GCN+GT \cite{veyseh2020improving}}:
     It employs the given aspect terms to customize the hidden vectors and benefits from the overall dependency-based importance scores of the words.
    \item \textbf{kumaGCN+BERT \cite{chen-etal-2020-inducing} }:
    considers latent graph structures for aspect sentiment classification by investigating a variety of neural networks for structure induction, and
    novel gated mechanisms to dynamically combine different structures.
    \item \textbf{MCRF-SA \cite{xu-etal-2020-aspect}}: 
    Utilizes syntactic dependency trees and attention mechanisms to model aspect-context interactions.
    \item \textbf{BiSyn-GAT+ \cite{liang-etal-2022-bisyn}}: 
    A bidirectional syntax-aware network that integrates syntactic constraints for sentiment prediction.
    \item \textbf{MGFN \cite{tang-etal-2022-affective}}: 
    Incorporates affective knowledge propagation with graph neural networks for sentiment enhancement.
    \item \textbf{NADS \cite{cao-etal-2022-aspect}}: 
    Leverages contrastive learning to distinguish aspect-specific representations from context.
    \item \textbf{SSEGCN \cite{zhang-etal-2022-ssegcn}}: 
    A syntax-enhanced graph convolutional network with selective position encoding.
    \item \textbf{dotGCN \cite{chen2022discrete}}: 
    Proposes discrete latent variables for modeling sentiment polarity transitions.
    \item \textbf{APARN \cite{ma-etal-2023-amr}}: 
    Integrates Abstract Meaning Representation (AMR) graphs with transformer architectures.
    \item \textbf{TF-BERT \cite{zhang-etal-2023-span}}: 
    A span-based extraction approach for joint aspect-sentiment detection.
    \item \textbf{SPT \cite{zhang-etal-2023-empirical}}: 
    Comprehensive empirical study comparing syntactic vs. semantic features in sentiment analysis.
    \item \textbf{A2SMvCL \cite{chai-etal-2023-aspect}}: 
    Multi-task framework with aspect-aware contrastive regularization.
    \item \textbf{AUG+DIS \cite{wang-etal-2023-reducing}}: 
    Focuses on reducing semantic gaps through knowledge distillation techniques.
    \item \textbf{SPT+BERT \cite{zhang-etal-2023-empirical}}:
    A method enables large-scale SPT corpus construction, complemented by comprehensive analysis of SPT techniques and experimental validation of their efficacy in advancing aspect-level sentiment understanding through ABSA tasks.
    \item \textbf{IDG \cite{li2024iterative}}: 
    Iterative refinement method using large language models for aspect-sentiment coherence.
    \item \textbf{ICL+LLM \cite{zhou2024comprehensive}}:
     Proposes three fine-tuning-free demonstration selection strategies that enhance LLMs' few-shot learning capabilities, enabling them to outperform fine-tuned Small Language Models (SLMs) and achieve new state-of-the-art performance.
    \item \textbf{GCNet+BERT \cite{zhou2024gcnet}}:
    A method that utilizes global information to guide context encoding has been constructed.
    \item \textbf{ME+LLM \cite{li2025exploring}}:
    A method that explores model editing(ME) as an efficient and interpretable fine-tuning method for LLM-based aspect-based sentiment.
 classification.
\end{itemize}

\begin{figure}[t]
    \centering
    \includegraphics[width=1.0\linewidth, trim=0 40 0 0]{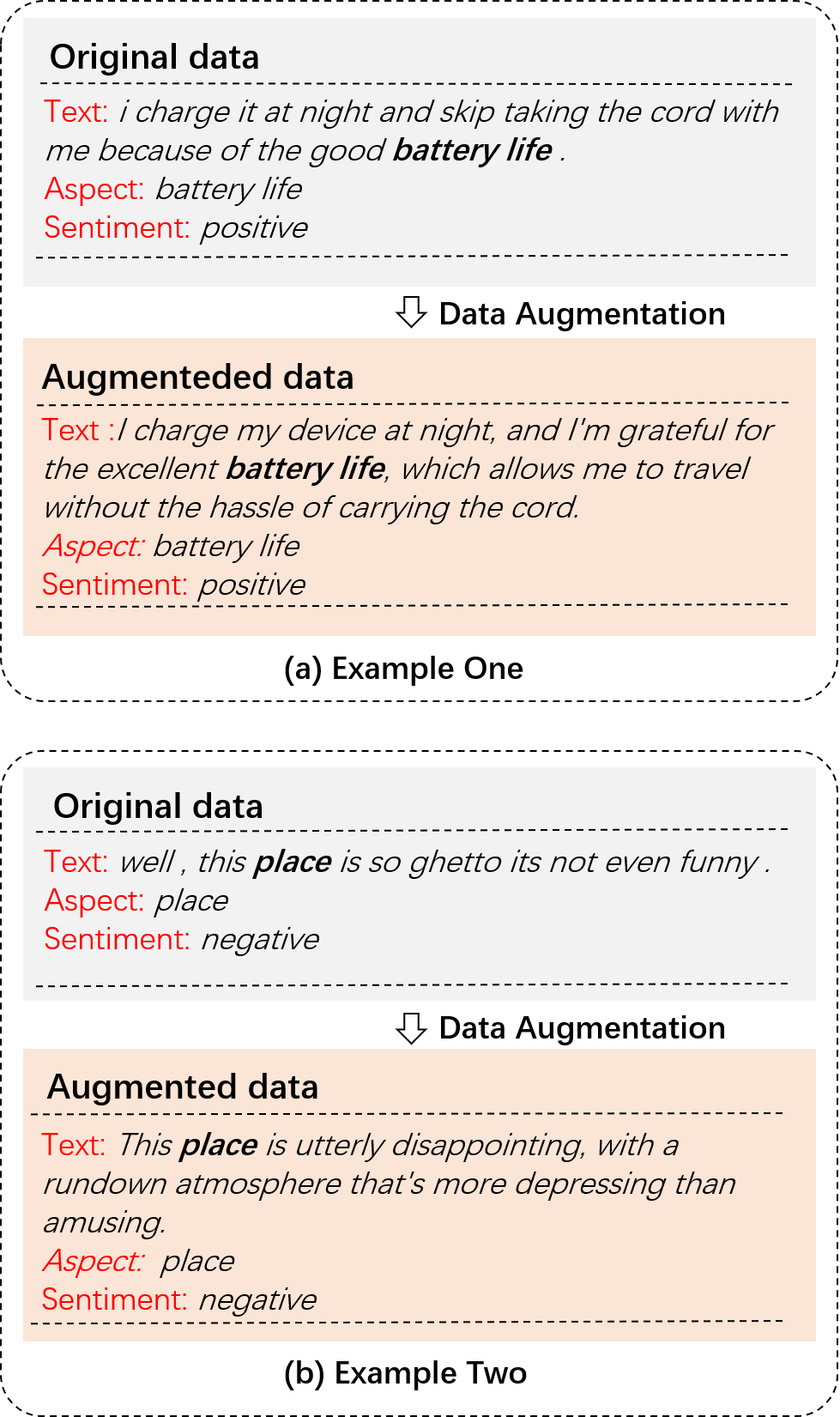}
    \caption{Some examples of enhanced data generated by the data augmentation model. The examples are extracted from the experimental training dataset, where the thick part of the table is the aspect items of the text.}
    \label{fig:data}
    \vspace{-0.3cm}
\end{figure}

\begin{table}[t]
\begin{center}
        \centering
        \scalebox{0.95}{
        \begin{tabular}{l l | r | r| r}
        \toprule
        & \textbf{Setting} & \multicolumn{1}{c|}{\textbf{Pos. \#}} & \multicolumn{1}{c|}{\textbf{Neu. \#}} & \multicolumn{1}{c}{\textbf{Neg. \#}} \\
        \midrule
        \multirow{2}{*}{\textbf{LAP14}} 
        & Standard & 1,988 & 928  & 1,740  \\
        & Balanced & 1,988 & 1,988 & 1,988 \\
        \midrule
        \multirow{2}{*}{\textbf{REST14}} 
        & Standard & 4,328 & 1,274 & 1,614 \\
        & Balanced & 4,328 & 4,328 & 4,328 \\
        \midrule
        \multirow{2}{*}{\textbf{REST15}} 
        & Standard & 1,814 & 72 & 508 \\
        & Balanced & 1,814 & 1,814 & 1,814\\
        \midrule
        \multirow{2}{*}{\textbf{REST16}} 
        & Standard & 2,458 &   138 &   874 \\
        & Balanced & 2,458 & 2,458 & 2,458\\
        \bottomrule
        \end{tabular}
        }
\end{center}
\vskip -0.1cm
\caption{\label{tab:enhanced dataset}
    The statistics of the final training dataset with different settings, where the number of instances with different sentiment polarities is reported.
}
 \vskip -0.5em       
\end{table}

\begin{table*}[t]
\centering
\begin{tabular*}{\linewidth}{@{\extracolsep{\fill}} l | c c | c c | c c | c c @{}}
\toprule
\multirow{2}{*}{\textbf{Models}} 
& \multicolumn{2}{c|}{\textbf{LAP14}} 
& \multicolumn{2}{c|}{\textbf{REST14}} 
& \multicolumn{2}{c|}{\textbf{REST15}}  
& \multicolumn{2}{c}{\textbf{REST16}} \\
\cmidrule(lr){2-3} \cmidrule(lr){4-5} \cmidrule(lr){6-7} \cmidrule(lr){8-9}
& ACC & F1 & ACC & F1 & ACC & F1 & ACC & F1 \\
\midrule
baseline
&67.24	&53.43	&77.41	&63.10	&68.10	&51.68	&78.76	&52.63 \\
only DA
&80.09	&76.09	&85.09	&75.87	&88.99	&75.99	&94.44	&81.30 \\
only DA*
&81.03	&79.04	&86.61	&80.80	&88.99	&78.52	&92.65	&81.94 \\
\midrule
DA+RL 
&82.60	&78.23	&87.23	&79.25	&91.23	&80.43	&93.63	&\textbf{81.59} \\
DA+RL* 
& \textbf{84.80} & \textbf{82.10} & \textbf{87.95} & \textbf{81.89} & \textbf{92.54} & \textbf{89.96} & \textbf{94.61} & 77.31 \\
\bottomrule
\end{tabular*}
\vspace{-0.2cm}
\caption{The performance comparison of our best model (using both DA and RL) and the model that only uses DA and the baseline that does not use DA. The models marked with ``*'' adopt the augmented data distribution of average labels in the data augmentation stage.}
\label{tab:baseline}
\vspace{-0.3em}
\end{table*}

\begin{table*}[t]
\centering
\begin{tabular*}{\linewidth}{@{\extracolsep{\fill}} l | c c | c c | c c | c c @{}}
\toprule
\multirow{2}{*}{\textbf{Models}} 
& \multicolumn{2}{c|}{\textbf{LAP14}} 
& \multicolumn{2}{c|}{\textbf{REST14}} 
& \multicolumn{2}{c|}{\textbf{REST15}}  
& \multicolumn{2}{c}{\textbf{REST16}} \\
\cmidrule(lr){2-3} \cmidrule(lr){4-5} \cmidrule(lr){6-7} \cmidrule(lr){8-9}
& ACC & F1 & ACC & F1 & ACC & F1 & ACC & F1 \\
\midrule

GCN+GT\cite{veyseh2020improving}       
& 82.8~~~& 80.2~~~& 87.2~~~& 82.5~~~& -     & -     & -     & -   \\

kumaGCN+BERT \cite{chen-etal-2020-inducing}                     
& 81.98 & 78.81 & 86.43 & 80.30 & 86.35 & 70.76 & 92.53 & 79.24     \\

MCRF-SA   \cite{xu-etal-2020-aspect} 
& 82.86 & 73.78 & 77.64 & 74.23 & 80.82 & 61.59 & 89.51 & 75.92 \\

BiSyn-GAT+  \cite{liang-etal-2022-bisyn} 
& 82.91 & 79.38 & 87.94 & 82.43 & - & - & - & - \\

MGFN  \cite{tang-etal-2022-affective} 
& 81.83 & 78.26 & 87.31 & 82.37 & - & - & - & - \\

NADS  \cite{cao-etal-2022-aspect} 
& 82.75 & 79.95 & 87.67 & \textbf{82.59} & - & - & - & - \\

SSEGCN  \cite{zhang-etal-2022-ssegcn} 
& 81.01 & 77.96 & 87.31 & 81.09 & - & - & - & - \\

dotGCN  \cite{chen2022discrete} 
& 81.03 & 78.10 & 86.16 & 80.49 & 85.24 & 72.74 & 93.18 & 82.32 \\

APARN  \cite{ma-etal-2023-amr} 
& 81.96 & 79.10 & 87.76 & 82.44 & - & - & - & - \\

TF-BERT  \cite{zhang-etal-2023-span} 
& 81.80 & 78.46 & 87.09 & 81.15 & - & - & - & - \\

SPT  \cite{zhang-etal-2023-empirical} 
& - & 78.68 & - & 81.59 & - & - & - & - \\

A2SMvCL  \cite{chai-etal-2023-aspect} 
& 82.12 & 78.82 & 87.86 & 82.41 & 86.74 & 75.05 & 93.42 & \textbf{83.80} \\

AUG+DIS  \cite{wang-etal-2023-reducing} 
& 81.56 & 75.92 & 86.37 & 80.63 & 83.98 & 70.86 & 91.45 & 78.12 \\

SPT+BERT\cite{zhang-etal-2023-empirical}
& - & 78.68  
& - & 81.59 
& - & -
& - & -
\\

*IDG \cite{li2024iterative} 
& 82.49 & 79.62 & 87.50 & 81.68 & 87.13 & 75.17 & 92.95 & 82.83 
\\

*ICL+LLM\cite{zhou2024comprehensive}
& 81.04 & -  
& 87.35 & - 
& 87.27 & -
& - & -
\\

GCNet+BERT \cite{zhou2024gcnet}
& 80.79 & 77.61 & 87.08 & 81.35 & - & - & - & - \\

*ME \cite{li2025exploring}
& 77.20 & -  
& - & - 
& - & -
& - & -
\\

\midrule
\textbf{*Our Best Model} 
& \textbf{84.80} & \textbf{82.10} & \textbf{87.95} & 81.89 & \textbf{92.54} & \textbf{89.96} & \textbf{94.61} & 77.31 \\
\bottomrule
\end{tabular*}
\caption{The comparison of our approach with previous state-of-the-art studies on all datasets for ABSA, where the approaches that use LLMs are marked by ``*''.}
\label{tab:sota}
\vspace{-0.2cm}
\end{table*}

\subsection{Data Augmentation Details}

The prompt template used by the data-enhanced model is shown in Table \ref{tab:my_label}. This template enables LLM instructions to follow and generate training data that meets the requirements of the ABSA task. Examples of the generated enhanced data and original data are shown in Figure \ref{fig:data}.

To evaluate the effect of our design for generating balanced training data, we try two settings.
The first ``\textbf{standard}'' setting generates one augmented training instance for each training instance in the original training set.
The second ``\textbf{balanced}'' setting produces balanced training data through the following process.
We firstly count the number of positive, neutral, and negative instances in the original training data and identify the largest class size. 
For example, on LAP14 the maximum class size is 994 (see Table \ref{tab: dataset}).
We then randomly sample from any class whose size is below this threshold, duplicating examples until each label reaches the largest class size, yielding a balanced original dataset.
For example, on LAP14, we extend the number of instances with neutral and negative sentiment polarities to 994.
Finally, we apply the same augmentation procedure as in the \textbf{standard}’’ setting to every instance in this balanced set, producing one synthetic example per original example and forming the balanced augmented set.
For both the ``\textbf{standard}'' and ``\textbf{balanced}'' settings, we merge the augmented samples with their corresponding original data (for the balanced condition, with the balanced original data). 
The final sizes and class distributions of the merged training sets are reported in Table \ref{tab:enhanced dataset}.

\subsection{Implementation Details}

A good text representation plays a crucial role in achieving promising model performance in NLP tasks \cite{mikolov2013efficient,song-etal-2017-learning,han2018hyperdoc2vec,ijcai2018-607,devlin-etal-2019-bert,song2021zen,grattafiori2024llama}.
In the experiments, we chose two widely used LLMs, namely Qwen-2.5 (1.5B)\footnote{We download the model from \url{https://huggingface.co/Qwen/Qwen2.5-1.5B-Instruct}.} \cite{yang2024qwen2} and LLAMA-3 (8B) \footnote{We download the model from \url{https://huggingface.co/meta-llama/Meta-Llama-3-8B-Instruct}.}  \cite{grattafiori2024llama}, where Qwen-2.5 (1.5B) is used as the ABSA model and LLAMA-3 (8B) is used for the data augmentation.
For both LLMs, we use the standard architectures and default hyperparameter settings. 
Specifically, the Qwen2.5 (1.5B) model has 28 layers of multi-head self-attention, and the vector dimension of each hidden layer is set to 1,536. 
LLAMA-3 (8B) has 32 hidden layers, and the vector dimension of each hidden layer is 4,096. 
For hyperparameters, we set the batch size to be 4 and the learning rate to be 1e-5.
For other hyperparameters, we tune them on the development set based on the model's performance.
Herein, since all datasets do not contain an official development set, we randomly select 10\% of the training data and use it as the development set to tune the hyperparameters.
Final models with the tuned parameters are trained on the entire training data and evaluated on the test data.
Following existing studies \cite{mao2019aspect,tang2020dependency,chen2022discrete,tian-etal-2023-end}, we evaluate all models through accuracy and F1 scores\footnote{We use the \texttt{Scikit-learn} package (\url{https://scikit-learn.org/stable/}) to compute the metrics.}.
All experiments are run on eight NVIDIA A40 GPUs with 48GB of memory.

\section{Results and Analysis}

\subsection{Overall Results}

The performance of our approach and the baselines are reported in Table \ref{tab:baseline}, where the approaches marked by ``$^*$'' generate augmented data that balances the label distribution of the final training data (i.e., the ``balanced'' setting).
There are the following observations.
First, the model using data augmentation (i.e., ``only DA'' and ``only DA$^*$'') performs better than the baseline model. 
This is because data augmentation increases the training data size, thus enabling the ABSA model to learn more contextual knowledge. 
Second, our data augmentation approach with reinforcement learning (i.e., ``DA+RL'' and ``DA+RL$^*$'') achieves better performance than the model that only uses data augmentation. 
The explanation is that the data augmentation LLM learns to generate high-quality augmented data that helps ABSA through the reinforcement learning process, which further improves the ABSA model that is trained on the augmented data.
This observation also highlights the value of the quality of the augmented data for the ABSA task.
Third, comparing approaches with and without using balanced training data (i.e., the ones with $^*$ and the ones without $^*$), we find that training the ABSA model with balanced label distributions improves the model performance, especially in the F1 score that reflects the classification effect of the model. 
This indicates that a uniform enhanced data label distribution reduces the overfitting risk of the model and improves the model performance. 

\begin{figure}
    \centering
    \includegraphics[width=1.0\linewidth, trim=0 30 0 0]{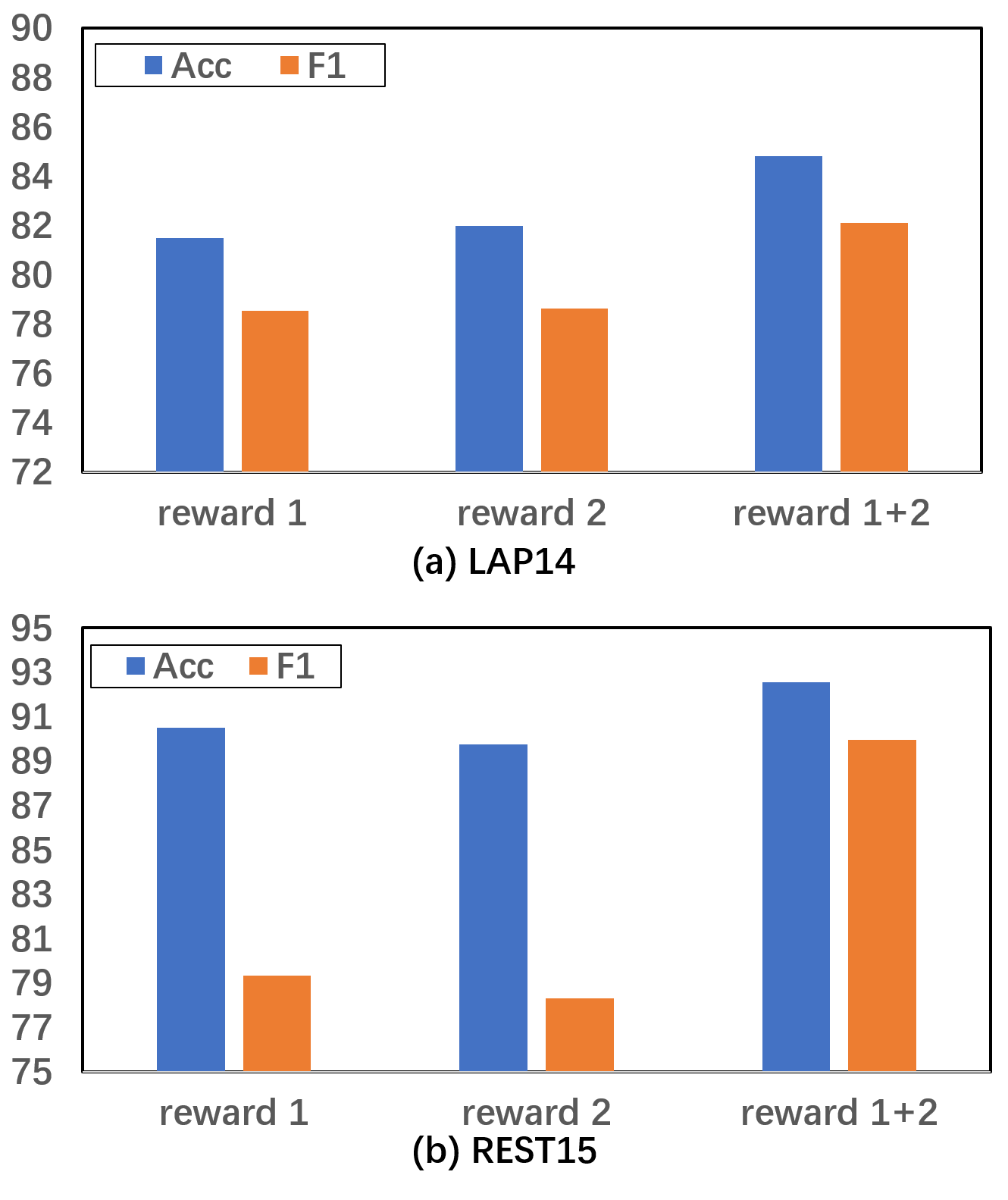}
    \caption{The experiment results after training with data augmentation using different reward functions in the LAP14 dataset and the REST15 dataset.}
    \label{fig:reward}
    \vspace{-0.2cm}
\end{figure}

We further compare our approach with existing studies, where the results are reported in Table \ref{tab:sota} (models that use LLMs are marked by ``*'').
Our model outperforms most existing studies, with noticeable improvements observed on the F1 score.
Particularly, compared with \citet{li2024iterative} that also utilizes an LLM to generate enhanced training data and filter the low-quality ones, 
we optimize the data augmentation LLM through DPO, which enables the LLM to generate high-quality data directly without quality filtering.

\begin{figure}
    \centering
    \includegraphics[width=1.0\linewidth, trim=0 30 0 0]{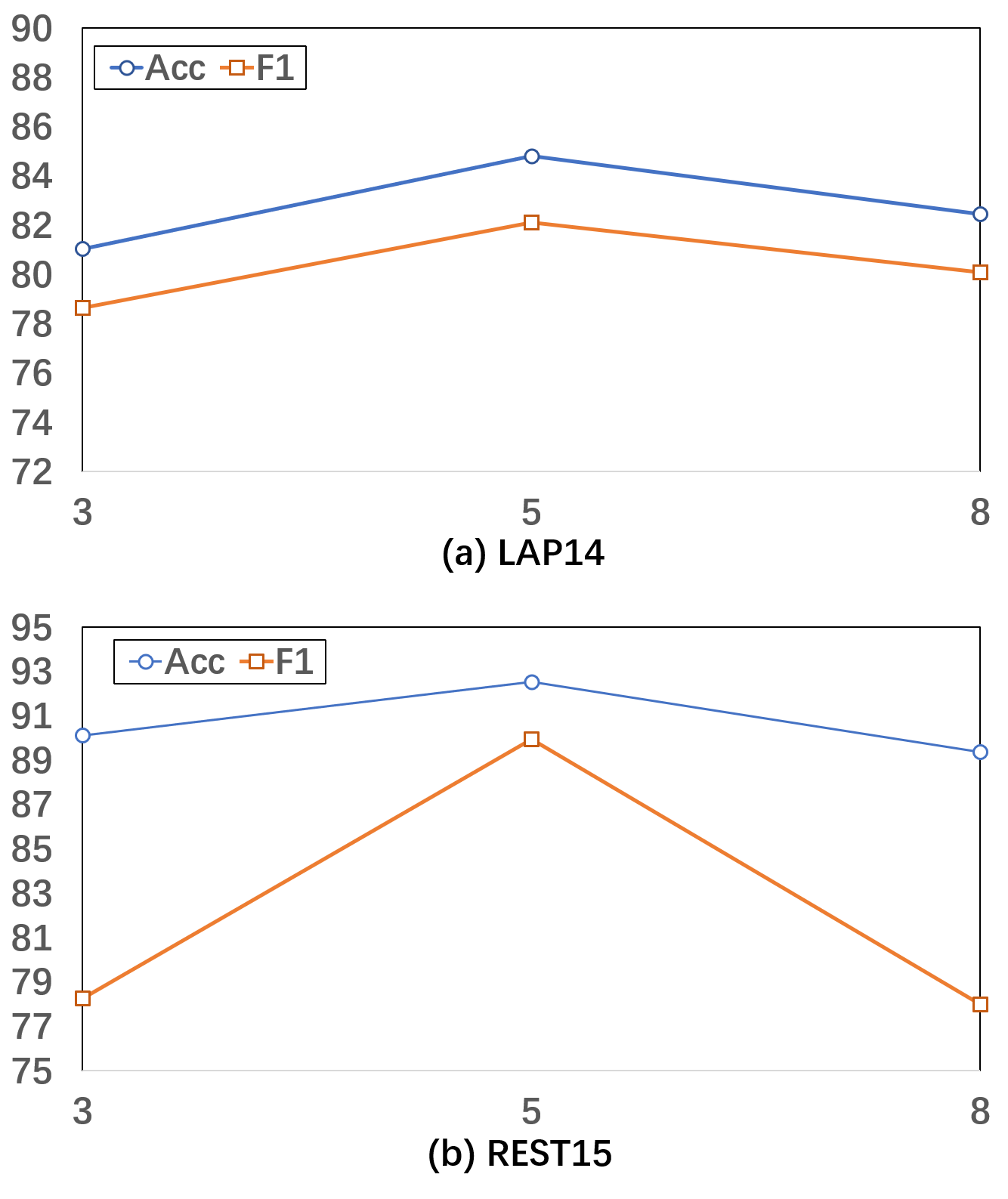}
    \caption{The influence of using LLM to generate different numbers of augmented data for preference dataset construction in LAP14 and REST15 datasets.}
    \label{fig:number}
    \vspace{-0.2cm}
\end{figure}

\subsection{Ablation Study}

We conduct ablation studies under two settings.
The first is to analyze the effect of different rewards and the second explores how the number of augmented instances to build the preference data for reinforcement learning influences the final ABSA model performance.
Figure \ref{fig:reward} shows the model performance when we only use the sentiment prediction function (i.e., ``reward 1'' ), only use the text relevance function (i.e., ``reward 2'' ) and combine the two reward functions (i.e., ``reward 1+2'') in the DPO stage.
Comparing approaches with and without using reward 1, we find that the model using the two reward functions outperforms the model that only uses the text relevance function.
The explanation is that, in the absence of ensuring the accuracy of the sentiment predicted by the generated augmented data, overly diverse generation may instead lead to a decline in the quality of the augmented data, thereby affecting the performance of the ABSA model.
Meanwhile, our model using the two reward functions also performs better than the model that only uses the sentiment prediction function (i.e., reward 2).
This indicates that if only the sentiment prediction rewards are used, the performance of the data augmentation ABSA model is not significantly improved because the generated text has poor diversity and does not enhance the context-learning ability of the model.

The results of the second setting are reported in Figure \ref{fig:number}, where we try values of 3, 5, and 8.
We find that when we only use three enhanced texts for preference data construction, the performance of the ABSA model is lower than that of our baseline model (i.e., ``5 enhanced texts'' ).
Meanwhile, we find that although the model constructed using eight enhanced texts for preference data outperforms the baseline model, the overall improvement is relatively small.
This indicates that the performance of the ABSA model increases with the increase of the number of texts generated by the data augmentation model when we build the preference dataset.  
However, when the number of generated texts reaches five, the improvement of the model performance tends to be stable. 
The explanation is that the generated texts of the LLM are obtained based on top-k sampling.  
When the number of different generated texts is large, lower-probability texts of poorer quality are adopted, which show less improvement on the ABSA task.

\begin{figure}[t]
    \centering
    \includegraphics[width=1.0\linewidth, trim=0 30 0 0]{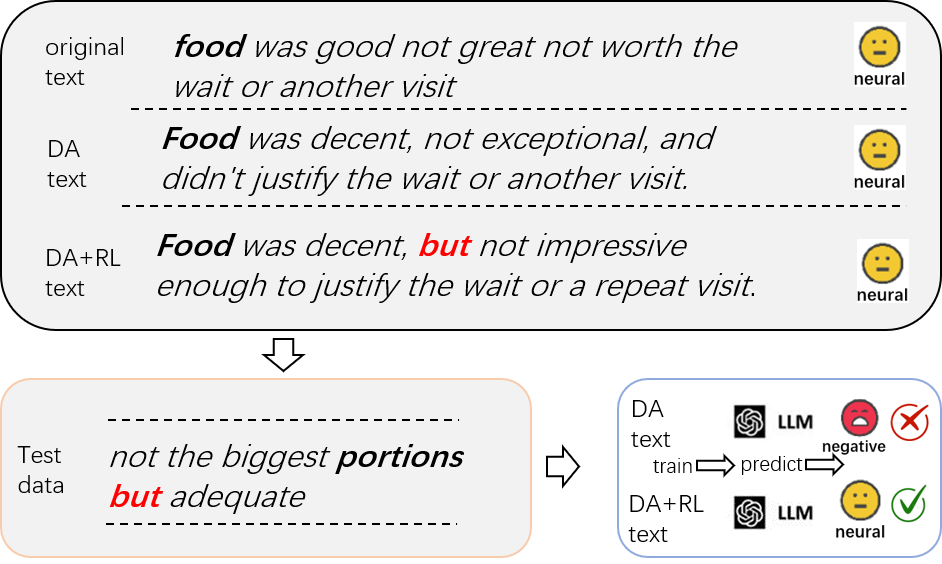}
    \caption{A case study on the data augmentation effect of reinforcement learning, where all the instances are data from the REST15 dataset. The aspect terms are highlighted in boldface, and similar transitional structures are represented in red colors.}
    \label{fig:case study}
    \vspace{-0.1cm}
\end{figure}

\subsection{Case Study}
To illustrate how our approach improves the ABSA task, we perform a case study and report the results in Figure \ref{fig:case study}.
In the figure, we present a test sentence and the aspect term, as well as the predictions from the baselines and our approach, where our approach makes the correct prediction while the baselines fail to do so.
We also present the augmented training instances produced by the ``only DA'' baseline (without using RL) and our approach (with RL).
As shown in the Figure, the augmented training instance generated by our approach shares some contextual information with the test instance, whereas the augmented training instance produced by the ``only DA'' baseline fails to do so.
This allows the ABSA model in our approach to learn the correspondence between the contextual information and the sentiment label from the augmented data and thus improves the model performance.
Overall, this case study demonstrates the effectiveness of our approach.

\section{Related Work}

\subsection{Aspect-based Sentiment Analysis}

Different from sentence-level sentiment analysis, ABSA focuses on determining the sentiment polarity of specific aspect words in a sentence, providing a more fine-grained and entity-oriented analysis \cite{li2018transformation,xu2019bert,chen-etal-2020-joint,qin-etal-2021-improving-federated,qin-2022-complementary,simmering2023large,tian-etal-2023-end}.
Therefore, it is essential to have a deep understanding of the text to perform ABSA well.
In recent years, researchers leverage pre-trained language models (e.g., BERT \cite{devlin-etal-2019-bert}) to effectively capture contextual information, particularly the essential tokens associated with the aspect term 
\cite{xue2018aspect,mao2019aspect,jiang2019challenge,tang2020dependency,wang2020relational}.
To further improve the modeling of the contextual information, there are studies that leverage different types of additional information to improve model performance.
Among them, many studies use syntactic information (e.g., dependency tree from off-the-shelf dependency parsers) in the running text \cite{tang2020dependency,tian2021aspect}, where graph-based approaches (e.g., graph convolutional networks (GCN) \cite{DBLP:journals/corr/KipfW16} and graph attention networks (GAT) \cite{Velickovic2017GraphAN}) are used to encode the syntactic information \cite{zhang2019aspect,wang2020relational,zhang2020convolution,zhang-etal-2022-ssegcn,zhang-etal-2023-span}.
Overall, these architectures aim to model the interaction between aspect terms and their surrounding context.
Moreover, LLMs are adapted to the complex aspect terms and context association required for precise sentiment classification in ABSA tasks due to their powerful context understanding capabilities, making it possible to leverage LLMs to improve ABSA \cite{simmering2023large,luo2024panosent,negi2024hybrid,ding2024boosting,tian2025large}.
Different from existing studies that design advanced architectures for better context modeling, our approach aims to generate more high-quality training data through data augmentation, so as to improve ABSA.

\subsection{Data Augmentation}

Data augmentation (DA) utilizes various approach to alter the structure of the original text to generate new data, with the aim of expanding the training dataset to alleviate the problem of data scarcity \cite{zhang2016characterlevelconvolutionalnetworkstext,xie2017datanoisingsmoothingneural,xie2020unsuperviseddataaugmentationconsistency}. 
Early work focuses on perturbations at the lexical level. 
For example, EDA (Easy Data Augmentation) \cite{wei-zou-2019-eda} generates different data through simple operations such as synonym replacement, random insertion, position exchange, and random deletion, but is limited by the rule system.
Problems such as semantic offset may occur in the enhanced samples. 
Although the approach at this stage increases the data scale, it faces the dual challenges of semantic fidelity and grammatical rationality.
To overcome the problem of semantic distortion, researchers turn to semantic-preserving augmentation. 
The Back Translation approach \cite{fan2021beyond} retains the core semantics through intermediate language translation and achieves an improvement in the accuracy of emotion prediction in the emotion classification task \cite{yu2018fast}, but its generation diversity is limited by the scale of parallel corpora. 
Only a few variations of sentence patterns are produced. 
Furthermore, models such as CBERT \cite{wu2019conditional} improve the consistency between the enhanced data and the original label through tag-aware mask prediction, taking into account both context semantics and sentiment polarity constraints simultaneously during the mask language modeling process.
However, the enhanced data often lacks sufficient contextual information. 
Therefore, the latest DA research focuses on context-aware enhancement driven by LLMs. 
The prompt-enhanced DA \cite{li2024iterative} design an iterative Prompt engineering framework and control the generation quality through a three-step prompt template.
First, specify the aspect term to be retained. 
Secondly, constrain the sentiment polarity. 
Finally, require sentence pattern diversity (such as ``\textit{Keep the sentiment polarity of the aspect term `service' unchanged and generate three sentence pattern variants}''), so as to improve the classification effect of the generated data on the ABSA task.
Compared with the existing approaches, in terms of improving the effect of data augmentation, our approach utilizes reinforcement learning to make up for the deficiencies of text generation diversity and quality of data augmentation in the ABSA field. 
By constructing two different rewards and using them to optimize the data augmentation model, our approach enhances the quality of the augmented data and thus improves the model performance.

\section{Conclusion}
In this paper, we propose a data augmentation approach enhanced by reinforcement learning, aiming to generate additional high-quality training data for ABSA tasks.
Specifically, we design a special prompt template to instruct the LLM to generate augmented text for ABSA tasks based on the original input text.
In enhancing the quality of the generated text, we further optimize the data augmentation LLM through DPO, where we construct a reward function to assess the quality of the augmented text and thereby establish a preference dataset for reinforcement learning.
We train the ABSA model using the combination of the original and the augmented data and evaluate our approach based on the performance of the ABSA model.
Extensive experiments on four English benchmark datasets for ABSA demonstrate the effectiveness of our approach, which outperforms strong baselines and most existing studies.

\bibliographystyle{acl_natbib}
\bibliography{tacl2021}

\end{document}